\documentclass{article}

\usepackage{lineno,hyperref}
\usepackage{makecell}
\usepackage{graphicx}
\usepackage{enumerate}
\usepackage{subfigure}
\usepackage{float}
\usepackage{setspace}
\usepackage{authblk}
\usepackage{abstract}

\title{Error Detection in a Large-Scale Lexical Taxonomy}
\author[a]{Sifan Liu}
\author[a]{Hongzhi Wang \thanks{corresponding author: wangzh@hit.edu.cn}}

\affil[a]{P.O.Box 750, Harbin Institute of Technology, Harbin, China}

\begin{document}
\begin{spacing}{1.0}

\maketitle

\begin{abstract}
	Knowledge base (KB) is an important aspect in artificial intelligence. One significant challenge faced by KB construction is that it contains many noises, which prevents its effective usage. Even though some KB cleansing algorithms have been proposed, they focus on the structure of the knowledge graph and neglect the relation between the concepts, which could be helpful to discover wrong relations in KB. Motived by this, we measure the relation of two concepts by the distance between their corresponding instances and detect errors within the intersection of the conflicting concept sets. For efficient and effective knowledge base cleansing, we first apply a distance-based Model to determine the conflicting concept sets using two different methods. Then, we propose and analyze several algorithms on how to detect and repairing the errors based on our model, where we use hash method for an efficient way to calculate distance. Experimental results demonstrate that the proposed approaches could cleanse the knowledge bases efficiently and effectively.
\end{abstract}
\section{Introduction}

Knowledge base (KB), such as YAGO~\cite{Weikum2007Yago}, and DBpedia~\cite{Yu2014DBpedia}, which typically contains a set of concepts, instances, and relations, becomes significantly important in both industry and academia~\cite{Nakai1992A}. With the increasing of data size on the web, KB automatic construction approaches~\cite{Murray1986Knowledge} are proposed to extract knowledge from massive data efficiently. Due to the low-quality of raw data and the limitation of KB construction approaches, automatically constructed KBs are nagged by quality problems, which will cause serious accidents in applications. Thus, KB cleansing approaches are in great demand to increase the usability of KBs.

Many KB and database cleansing algorithms have been proposed in recent years such as building directed acyclic graph and applying other new knowledge. The main idea of building directed acyclic graph method is to enumerate cycles and eliminate the relation with low trustworthy because cycles are highly likely to contain wrong relations. However, such approach focuses on the structure and level of the KB with the perspective of graph, and does not have the ability to detect low frequency errors which are not included in a cycle. As for applying other new knowledge such as another KB or internet knowledge, it would take lots of time and each KB contains its unique relations that would be very different from the one that we are dealing with.

In summary, these algorithms focus on the structure of the knowledge graph and external knowledge without sufficient usage of the relations in the KB that need to be cleansed. When we think about concepts, we usually consider the relation of two concepts. For example, bird and fish are two conflicting concepts while bird and animal are two related concepts. Motivated by this, we attempt to adopt the relations between instances sets corresponding to the concepts for effective KB cleansing. Such approach brings following technical challenges.

\begin{enumerate}[(1)]
	\item First, in a KB, the frequency of many relations is 1. For example, in Probase, 7M relations only emergence once. Thus, we can hardly derive error from frequency and have to seek frequency-independent error detection approaches.
	\item Second, current KBs are almost in large scale, which leads to inefficient data accessing during KB cleansing. It requires sophisticated data structures and algorithms for cleansing algorithms to achieve high performance.
	\item Third, with the existence of homonyms, even though a conflicting is discovered, it should be distinguished whether it is caused by wrong triples or homonyms. Such distinguishing is crucial for cleansing strategy determination and in great demand.
\end{enumerate}

Facing these challenges, we attempt to develop efficient KB cleansing algorithms in this paper. We observe that relation between concept sets corresponding to conflicting concepts could be adopted for KB cleansing.

Consider the example for motivation, the concepts of bird and fish are conflicting, which means \emph{a creature} could not be both a \emph{fish} and a \emph{bird}. Therefore, the wrong triple \emph{(eagle, Isa, fish)} in Probase is detected since \emph{eagle} is in the intersection of the concept sets corresponding to \emph{fish} and \emph{bird}.

Based on this observation, we develop two error detection algorithms with various distance measure between concept sets from different aspects. For effectiveness issues, these algorithms adopt Hamming distance and Jaccard distance respectively to take advantages of more trustworthy relations and take full use of the total domain size. These two algorithms could allow us to take full use of the information provided in Knowledge base itself. All these algorithms are based on set distance computation. For efficiency issues, the optimization of set distance computation is easier than that of large graph since complex structural information are not required to be collected and existing similarity join method for sets could be applied. To accelerate processing, we apply Simhash and Minhash LSH(Locality Sensitive Hashing for these algorithms respectively.

For detecting wrong triples, we develop two-level repair mechanism for a more accurate result. For the easy cases, we repair the triples according to frequencies. When the frequency-based approach does not work, we develop the crowdsourcing-based repair approach to distinguish homonyms.

The contributions of this paper are summarized as follows:

\begin{enumerate}[(1)]
	\item First, we detect errors in KBs according to the relation between the concept sets. Such approach makes full usage of the Isa relationship between concepts and is easy to accelerate for set operations.
	\item Second, we develop three efficient KB error detection algorithms for various distances with acceleration strategies as well as the two-level repair algorithm. Such KB error detection and repairing algorithms achieve universal by diversification.
	\item Third, experimental results demonstrate that the proposed algorithms outperform existing approaches and could cleanse the KB efficiently and effectively.
\end{enumerate}

%------------------------------------------------

\section{Distance-based Models}

Even though some KB error detection methods have been proposed, they fail to fully use of the relations in KB. Major KB error detection approaches could be classified into two kinds.

One is frequency-based approach~\cite{Van2005Data}. The low frequency means that the relation is seldom noticed in the corpus or the web. Therefore, some approaches have been proposed to use frequency to determine whether a relation is correct or not. Clearly, many relations with low frequencies are correct. In Table (a), we list the percentage of frequencies in different range. We randomly selected 100 relations in the Probase~\cite{DBLPconf/aaai/LiangXZHW17} to check the correctness, and the answer is given by human judgment. Andwe list the correctness rate of these data with different frequencies in the Table (b). From this experiment, just using frequency to determine whether a relation is wrong could be an unwise way to find errors and would achieve a low accuracy. We give specific example to show its limits.

\textbf{Example 1:} \emph{The frequencies of triples (\texttt{snake}, Isa, \texttt{herb}) and (\texttt{fruit fly}, Isa, \texttt{creature}) are both 1 in Probase. However, the latter is correct while the former is wrong. It shows that just using frequency fails to detect many errors.}

The other is structure-based approaches~\cite{Clauset2008Hierarchical,Gupte2011Finding,DBLPconf/aaai/LiangXZHW17}. It has been observed that most cycles in a knowledge graph contain wrong Isa relations since a prefect knowledge base should be acyclic. Therefore, eliminating cycles could correct wrong Isa relationships. However, many wrong Isa relations are not contained in any cycle since cycle is just a specific structure and many wrong errors do not happen to be in such specific structure.

\begin{table}[h]
	\centering
	\subtable[Percentage of Frequency]{
		\begin{tabular}{|c|c|}
			\hline
			Frequency & Percentage(\%) \\
			\hline
			1 & 65.3 \\
			$>1$ & 34.7\\
			\hline
		\end{tabular}
		\label{tab:PF}
	}
	\qquad
	\subtable[Accuracy of Frequency]{
		\begin{tabular}{|c|c|}
			\hline
			Frequency & Accuracy(\%) \\
			\hline
			1 & 65.0 \\
			$>1000$ & 100.0\\
			\hline
		\end{tabular}
		\label{tab:AF}
	}
\end{table}

Therefore, we attempt to find a more general approach for error detection. At first, we discuss the motivation of our distance-based approaches. Then we discuss the definition of distances as well as their combinations.

\subsection{Motivations}
We observe that many errors are caused by misclassification of instances to concepts. Using the example stated before, \emph{fish} and \emph{bird} are two obvious conflicting concept sets because there is no such \emph{creature} in the world is both a\emph{ bird} and a \emph{fish}. If the same \emph{creature} belongs to the concept of both \emph{fish} and \emph{bird}, at least one of these two Isa relations are wrong. According to this intuition, we detect errors using the intersection of conflicting sets.

We first define some related concepts.

\textbf{Definition 1 (Concept Sets)} \emph{Given a Knowledge base with Isa Relation, the weighted concept sets constructed based on this KB contains all the instances which have Isa relation with the concepts, and each instance is associated with a weight w(p).}

For example, we use the instance stated before, \emph{fish} is a concept in Porbase and we build a concept set called \emph{fish}, which contains all the instances such as \emph{tuna}, \emph{ salmon} and \emph{catfish}. And if we use the frequency as the weight w(p), the weight of these three are 1892, 3733 and 562.

\textbf{Definition 2 (Conflicting Concept Sets)} \emph{Given all the concept sets in a Knowledge base with Isa Relation. Consider two of these concept sets each, if these two concepts are incompatible in semantics, their corresponding concept sets are called conflicting concept sets.}

For example, we could construct concept sets \emph{bird} and \emph{fish} from the knowledge base according to Definition 1, it is obvious that \emph{bird} and \emph{fish} are conflicting concept sets since they are incompatible in semantics.

\begin{figure}[h]
	\centering
	\subfigure[Similar Sets 1]{
		\includegraphics[width=1.2in]{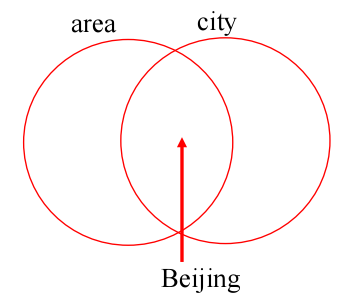}}
	\qquad
	\subfigure[Similar Sets 2]{
		\includegraphics[width=1.2in]{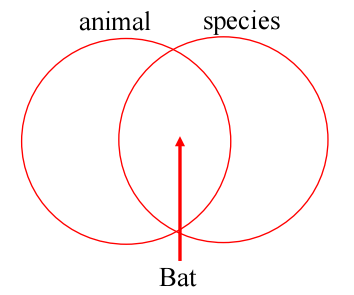}}
	\caption{Similar Concept Sets}
	\label{fig:redfig}
\end{figure}

\begin{figure}[h]
	\centering
	\subfigure[Conflicting Sets 1]{
		\includegraphics[width=1.2in]{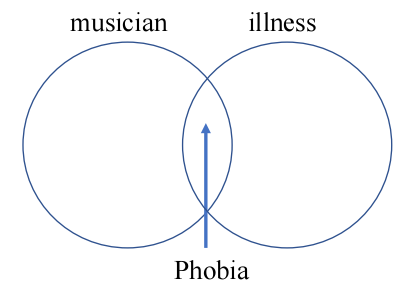}}
	\qquad
	\subfigure[Conflicting Sets 2]{
		\includegraphics[width=1.2in]{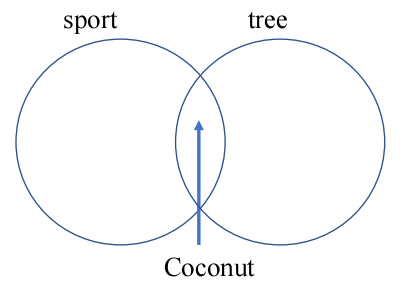}}
	\caption{Conflicting Concept Sets}
	\label{fig:bluefig}
\end{figure}

We now show how to use the intersection of two Concept Sets to detect error. In the figure showed above, we present two kinds of relations of concept sets in the knowledge base, Figure  \ref{fig:redfig} shows two similar concept sets, which means each set has some similar properties with the other, while Figure  \ref{fig:bluefig} shows two conflicting concept in semantics, which means the Conflicting Concept Sets stated before. It is obvious that the intersection of two Conflicting Sets contains errors of misclassification of instances. Therefore, computing the intersection of conflicting concept sets allows us to detect errors of low frequency with high accuracy. Clearly, the ideal intersection of two conflicting concept sets should be empty. On the other hand, if their intersection is not empty, errors have a highly probability to occur on the instance in their intersection. In this way, we perform error detection with set operations, which are pretty efficient. Additionally, this approach are not affected by the frequencies.

For a human, it is easy to distinguish conflicting concepts. However, the challenge is to find them automatically in massive knowledge bases. Inspired by the similarity measure for documents~\cite{Tong2014Document}, we attempt to use the distance between concept sets to determine conflicting sets. Intuitively, with the assumption that a KB is large enough to contain all instances of each concept, if two concept sets are very different in their instances, they are different in semantics correspondingly. Based on this intuition, we determine the conflicting concept sets according to the distance. Thus, the definition of distance is crucial to conflicting concept sets determination, and we will discuss it in the next subsection.

\subsection{Distances}

Many distances or similarities have been proposed to measure the difference of sets. In order to fully use the information that provided by the KB, the more trustworthy relations should be considered importantly according to each concept sets, and the Hamming distance could take consideration of the importance of each relation. Besides, as we stated before, an ideal intersection of two conflicting concept sets should be empty. Therefore, we also need to consider the number of relations in the intersection, that is the reason to use Jaccard distance. In conclusion, we select two different distances for conflicting concept detection, Hamming distance~\cite{Hamming2014Error} and Jaccard distance~\cite{Jaccard2010THE}.

There definitions are listed as follows.

\emph{Hamming distance}: For two sets $A$ and $B$, we define a hash function $h(.)$ to map a set to a string. Thus, the Hamming distance between $A$ and $B$ $H(A,B)=h(A)  XOR  h(B)$.

\emph{Jaccard distance}: For two sets $A$ and $B$, their Jaccard distance $J(A,B)= 1-\frac{\left| A\bigcap B\right|}{\left| A\bigcup B\right|}$.

The main purpose to use these two distances is that it could serve as different criteria for us to determine the conflicting concept sets. The Hamming distance could calculate the distance between two hash signatures, and it could be easy for us to embed frequency of instance to the distance measure. The Jaccard distance uses the percentage of the number of intersection of two concept sets to determine the conflicting between two concept sets. And this is similar to the people cognition strategy since we know that the two conflicting concept sets have no instances in their intersection. At the same time, Jaccard could take use of the full domain size of these two concept sizes. Therefore, these two distances could work well to fully consider the information provided by the knowledge base and to detect errors.

\subsection{The combination of distances}
Therefore, as listed before, each of these distances has its own pros and cons. To achieve high accuracy, a feasible approach is to combine them to avoid the limitations. According to above discussions, we have the following combination strategies.

\begin{enumerate}[(1)]
	\item Using Hamming distance to generate the set $S_H$ of conflicting concept sets.
	\item Using Jaccard distance to generate the set $S_J$ of conflicting concept sets.
	\item Combining the result $S_H$ and result $S_J$ to get final result $S$.
\end{enumerate}

%------------------------------------------------

\section{Detecting Algorithm}

In the previous section, we show the approach of distance-based error detection. Due to the large scale of KBs, efficient and scalable algorithms are in demand. Thus, in this section, we develop efficient distance-based error detection algorithms.

For efficiency issues, we adopt hash method for the acceleration. That is, for each concept set $S$, we generate a signature $S_c$. Thus, a set $S$ is generated for all signatures. Then a distance join operation, which is similar to the similarity join, is performed on $S$ to generate the results.

For these three distances, we apply different hash function and distance join strategy for efficiency issues.

\textbf{Simhash Method for Hamming distance:}
Simhash is a fast algorithm for us to calculate Hamming distance in large scale data. However, one significant characteristic of this method is that it makes use of the weights of every instance, so the hash signature of each concept set is mainly depending on the larger weighted instance, and the smaller instance relation only act as a very small influence, therefore, it is useful to distinguish the conflicting when the two concept sets are very different with each other, while Hamming distance could not distinguish when two concept sets have very different domain size.

\textbf{Minhash LSH Method for Jaccard distance:} Minhash LSH is also a very useful method in dealing with the document duplicate removal or web page comparing, and it based on the Jaccard distance, the main advantages of Minhash LSH is that it could produce the Jaccard distance in a very short time, therefore, it is suitable for gigantic data size. However, as stated before, Jaccard distance faces two problems: the subset inclusion relations problem and the big set domain size influence.

\textbf{Combination of Simhash Method and Minhash LSH Method:} In order to find more conflicting concept sets with high accuracy, we propose to combine the Simhash Method and Minhash LSH Method stated above. Simhash Method could make full use of the high frequency information according to each relations because higher frequency represents higher trustworthy. Therefore, Simhash Method could find conflicting concept sets which have very little same high frequency relations in both of the concept sets, but it does not consider the differences between domain size and the influence of low frequency relations. However, the Minhash LSH Method makes full use of domain size and regard the each relation with the same importance without considering of frequency. Therefore, both Simhash Method and Minhash LSH Method could find conflicting concept sets with high accuracy but each of them may ignore some kind of conflicting while the other could detect just as stated above. And we will discuss our two-level repair method in the next section.

Therefore, to make a better detection, we propose to use the Combination of Simhash Method and Minhash LSH Method for general Knowledge base. And if we are detecting a Knowledge base without frequency, we propose to use Minhash LSH Method.

%------------------------------------------------

\section{Repairing Algorithm}

After computing the conflicting concept sets, we could achieve our main purpose of KB cleansing by computing the intersection of conflicting concept sets. Our main purpose is to repair errors, and there are another two kinds of relations that can be detected in the intersection of conflicting concept sets.

\textbf{Definition 3 (Errors)} \emph{ Given all the relations in an intersection of two conflicting concept sets, if there is an instance P with large differences between two weights regarding two different concept sets, the larger weight is bigger than B, and the low weight is smaller than L, then the relationship R with the smaller weight according to instance P is the error.}

For Examples, consider the intersection of \emph{bird} and \emph{fish} concept sets as we discussed before, the frequency of \emph{turekey} Isa \emph{bird} in Probase is 211, which means there are 211 sentences containing this Isa relation. And the frequency of \emph{turkey} Isa \emph{fish} in Probase is 1. It is obvious that turkey belongs to bird not fish. Therefore, we could say that the relationship turkey Isa fish is an error.

Since the concept sets we are dealing with is conflicting with each other, thus, for the second level of repairing, we could use the frequency in the KB as the weight to decide which relations should be deleted. In the same time we could also apply different weights for cleansing according to different situations.

\textbf{Definition 4 (Homonyms)} \emph{ Given all the relations in the intersection of two conflicting concept sets, if there is an instance P with both large weights, where the weights are both bigger than B, then the relations in both concept sets can be correct, and P is a homonym instance in both concept sets.}

As we stated before, the relationships in the intersection of two conflicting concept sets can be both correct, because one instance could have multiple meanings. But in the experiment, we find that conflicting concept sets could have very little homonyms because the similarity degree of these two sets is significantly low. Therefore, we propose the idea to give these instances sub-attributions to identify them in the different relations and concepts.

\textbf{Definition 5 (Suspicious relations)} \emph{ Given all the relations in the intersection of two conflicting concept sets, if there is an instance P with both very low weights regarding the both concept sets, where the weights are both lower than L, then the relations in both concept sets are suspicious.}

We still consider the intersection of \emph{bird} and \emph{fish} conflicting concept sets, the frequency of \emph{maple} Isa \emph{fish} is 1, and the frequency of \emph{maple} Isa \emph{bird} is also 1. As human, we could successfully judge these two relations to be both errors but if we look at the intersection of \emph{fish} and \emph{herb}, the frequency of \emph{ health supplement} Isa \emph{fish} is 1, and the frequency of \emph{ health supplement} Isa \emph{herb} is also 1. As human, we view these two relations differently according to different people, therefore, it is hard for automatically distinguish these suspicious relations.

we notice that the suspicious instances are usually both seldom using relations or even wrong relations in the conflicting concept sets. And it is even hard for human to raise up an agreement of right or wrong of these relations. Our cognition could provide very different conclusions. Therefore, there is no efficient automatically way to efficient classify these suspicious relations. In the experiment, we propose to build a suspicious knowledge base(SUSKB) and put these suspicious relations into SUSKB, and people could manually remove these relations in the SUSKB as they want.

%------------------------------------------------
\section{Experiment}

To verify the effectiveness and efficiency of the proposed approaches, we evaluate our approach in this section. We first applying our method on Probase and compare it with the latest error detection algorithms in precision. Next, we analyze the influence of the parameters based on 100 concept sets. And we finally show our method could also be used in other knowledge bases.

\subsection{Exp1: Applying on Probase}

In this experiment, we apply our methods on the core version of IsA data mined from billions of web pages, contains 5,376,526 unique concepts, 12,501,527 unique instances, and 85,101,174 IsA relations, and most of them have small frequencies as stated before.

Because we do not know the number of errors that contained in Probase, we then use the precision rate to evaluate our models, and we use the highest precision rate of the latest error detection method as comparison. Precision rate means the proportion of the truly wrong IsA relations in all relations detected by our methods. We randomly pick 500 wrong IsA relations to determine whether it is right or wrong. Since this work could only be done by people, we ask 50 volunteers to judge our results. And the final results are in Table \ref{tab:Data}

\begin{table}[h]
	\centering
	\scalebox{1.0}{
		\begin{tabular}{|c|c|c|c|c|c|}
			\hline
			\thead{Model} & \thead{Errors} & \thead{Concepts Set \\Precision(\%)} & \thead{Errors \\Precision(\%)}\\
			\hline
			\thead{baseline} & 74.2K & - & 91.3 \\
			\thead{Hamming Distance} & 100K & 83.3 & 89.2  \\
			\thead{Jaccard Distance} & 90K & 86.0 & 91.4 \\
			\thead{Combination} & 120K & 92.7 & 92.3 \\
			\hline
	\end{tabular}}
	\caption{Experimental Results}
	\label{tab:Data}
\end{table}

We can tell from the result that using the Combination distance could perform the best result. Because we apply hash method to conducting our results, our method largely depends on the parameters that we select. In the next subsection, we are going to analyze the parameters carefully using 100 concept sets randomly selected from the total concept sets.

\subsubsection{ Exp2: Analyzing for Parameters}

From the total results above, we know that Jaccard distance could bring a wonderful result to cleanse the knowledge bases. Since the Minhash LSH method that we use involves two parameters, one is the buckets number and the other is the threshold to determine whether two sets are conflicting with each other. We use 100 concept sets to analyze the influence of these parameters.

\begin{figure}[h]
	\centering
	\includegraphics[width=3.5in]{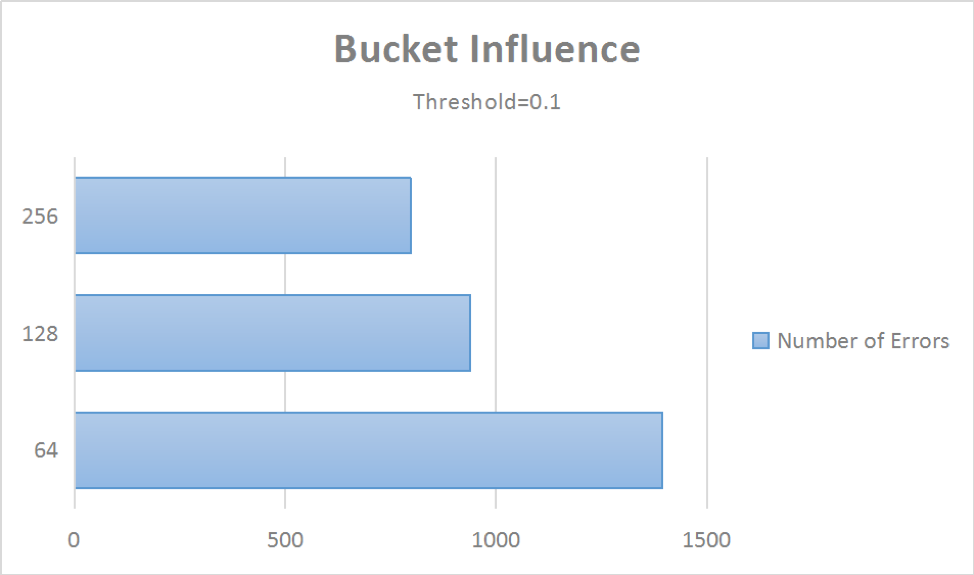}
	\caption{Bucket Influence}
	\label{fig:exp1}
\end{figure}

The Figure \ref{fig:exp1} shows the number of errors that can be found when we set the bucket number as 64, 128 and 256. It is easy to find that when the bucket number is 64, the errors of these 100 concept sets are most while the bucket number is 256 gets the lowest error number. This is in line with the algorithm of Minhash LSH. Since the number of bucket decides the possibility that the pair will be mapped into the same bucket. The smaller bucket number means a looser criteria to determine two concept sets are conflicting. Let's use a more intuitive way to analyze.
Consider five identical concept sets: animal, bird, fish, fruit, herb. And the result of the human judgement and the bucket number influence is showing below.

\begin{table}[h]
	\centering\scalebox{0.9}{
		\begin{tabular}{|c|c|c|c|c|}
			\hline
			\thead{Intersection} & \thead{Human\\Judgement} & \thead{64} & \thead{128} & \thead{256}\\
			\hline
			animal\_fruit & yes & no  & no & no \\
			animal\_herb & yes & yes & yes & yes \\
			animal\_bird & no & no  & no & no \\
			animal\_fish & no & no & no & no \\
			fruit\_herb & yes & no & no & no \\
			fruit\_bird & yes & yes & yes & yes \\
			bird\_fish & yes & yes & yes & yes \\
			fish\_fruit & yes & yes & no & no \\
			bird\_herb & yes & yes & no & no \\
			fish\_herb & yes & yes & no & no \\
			\hline
	\end{tabular}}
	\caption{Case Study}
	\label{tab:CaseStudy}
\end{table}

We can see clearly that when the bucket number = 64, there are six concept sets being determined to be conflicting while when it comes to 128 and 256, there is only four concept sets are conflicting with each other. Besides, there is one more important thing is that, as we can tell from the table, no matter how many conflicting concept sets are found, the judgment by the computer is right according to the human judgment. Therefore, if there need to find more errors from the knowledge base, we suggest setting a smaller bucket number, while if the criteria need to be strict, then the bucket number should be higher. In the following analysis of parameters, we choose to use 128.

Next, we are going to show the threshold influence. We know that the Minhash LSH is Locality Sensitive Hashing, which is often using to find the similarity in large scale data with high dimension. Using LSH is faster than linear search. Hamming distance and Jaccard distance in our method are all locality sensitive. Bellowing is the result when we change the threshold of the Minhash.

\begin{figure}[h]
	\centering
	\includegraphics[width=3.5in]{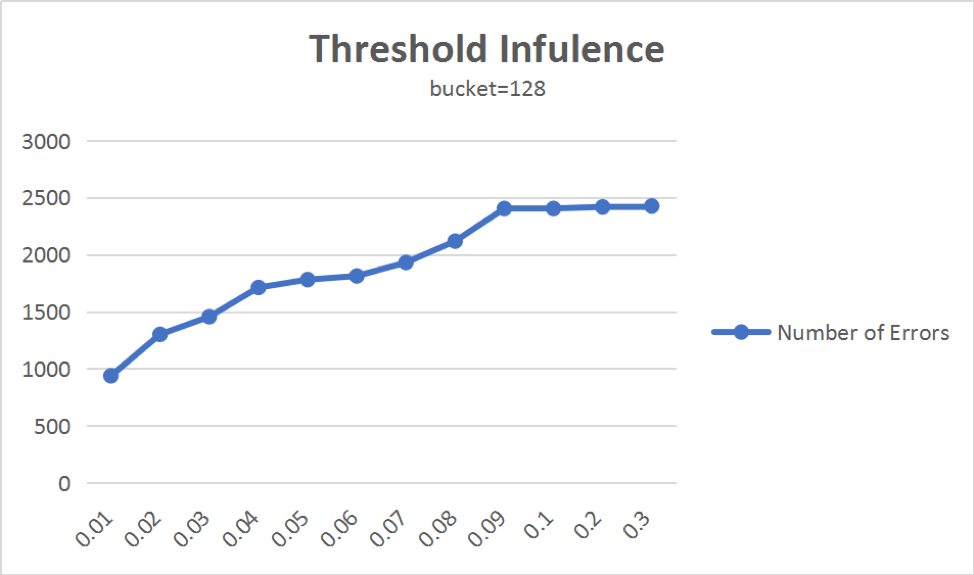}
	\caption{Threshold Influence}
	\label{fig:exp2}
\end{figure}

We can see clearly when the threshold is larger than 0.1, the number of the errors we find from the 100 concept sets doesn't change much, which means that the conflicting concept sets selected from the 100 concept sets are almost the same.

One of the important aspect of our method is that we use a two-step verification to ensure that the errors is more trustworthy. In the next step after selecting the conflicting concept sets using Hamming distance and Jaccard distance, we use three weight differentials to determine whether the relations in the intersection of two conflicting concept sets are errors, Homonyms, or Suspicious relations as we stated before. In the following, we pay attention to the weight differential to find errors, which means the differential needs to be large enough. We first show the error distribution when using different weight differentials.

\begin{figure}[h]
	\centering
	\includegraphics[width=3.5in]{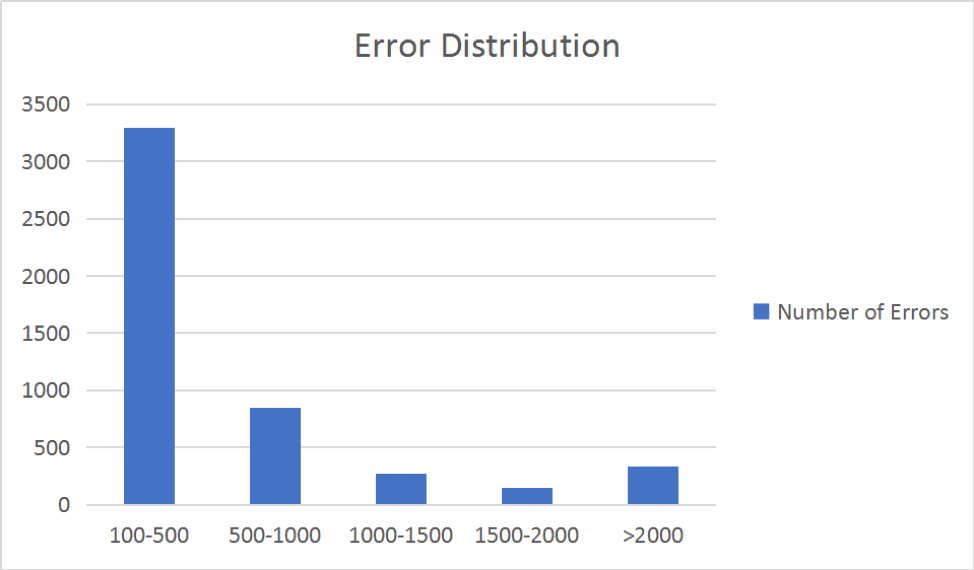}
	\caption{Error Distribution }
	\label{fig:exp3}
\end{figure}

From the error distribution, we can easily tell that the number of errors are decreasing as the differentials are setting stricter, which is obvious since the percentage of large frequency in Probase is small. Then We list all the results when selecting the different weight differential and different minimum weight. The results contain numbers of errors that can be found and the precision deciding by human volunteers. In our experiment, we make the bucket as 128 and the threshold as 0.01.

We set the minimum weights to be 1 to 10 and above 10 in different section of weights differential. Then, we measure the number of errors that can be found in 100 concept sets according to each situation, and we also show the errors truly wrong by human judgment. From the above result, we see that when we set the weights differential as 100 to 500 and the minimum weights to be 1, the number of errors is the most. And the minimum weights as 1 takes the largest proportion in each weights differential. Since our purpose is to find and remove errors, we need to obtain the highest precision rate. From the human judgment, we find that when the weights differential becomes higher, the precision rate is higher too. And when the minimum weight is less than 5, the precision rate remains high. However, when the minimum weight is larger or equal to 5, the precision rate drops to nearly 50.

\begin{figure}[H]
	\centering
	\subfigure[difference = 100 - 500]{
		\includegraphics[width=2.1in]{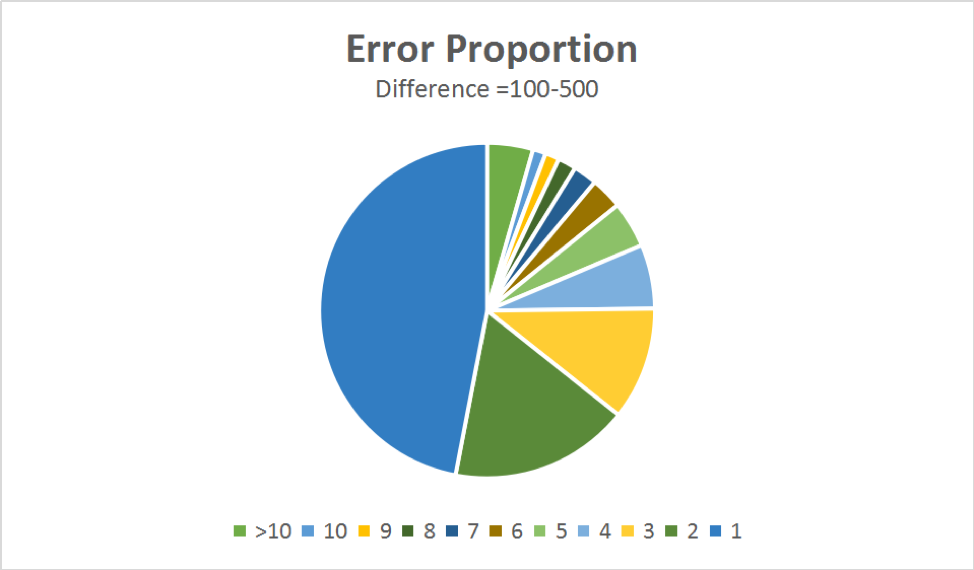}}
	\qquad
	\subfigure[difference = 100 - 500]{
		\includegraphics[width=2.1in]{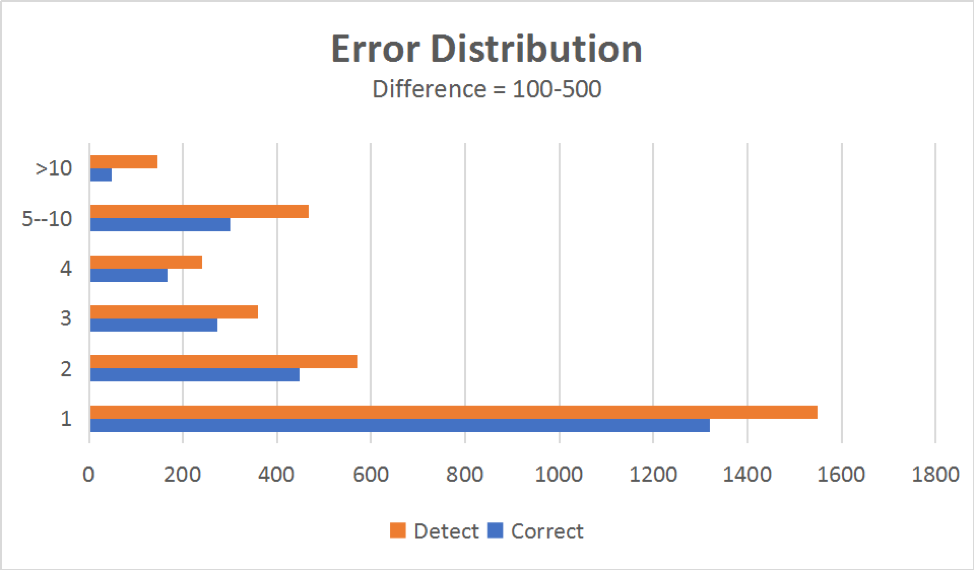}}
	\newline
	\subfigure[difference = 500 -1000]{
		\includegraphics[width=2.1in]{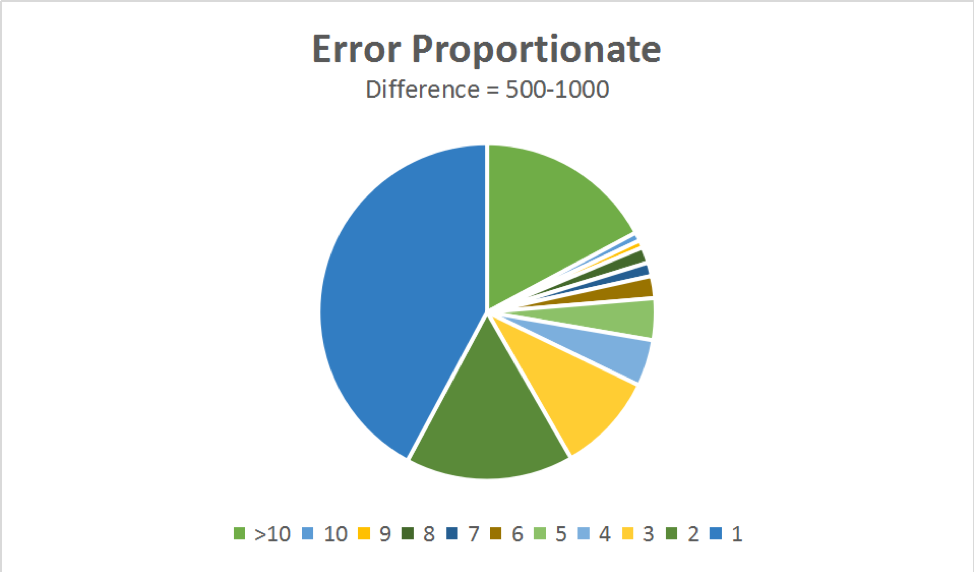}}
	\qquad
	\subfigure[difference = 500 -1000]{
		\includegraphics[width=2.1in]{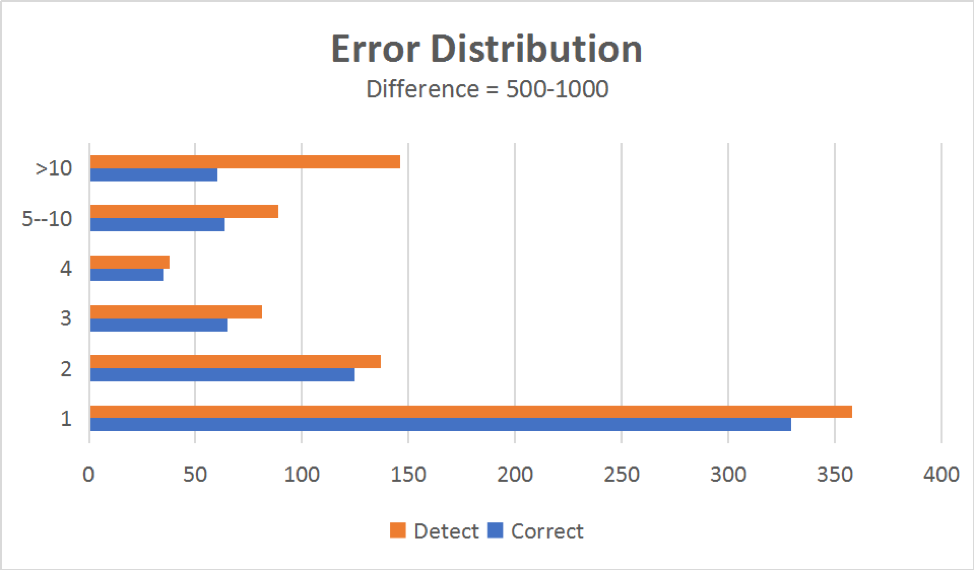}}
	\newline
	\subfigure[difference = 1000 -1500]{
		\includegraphics[width=2.1in]{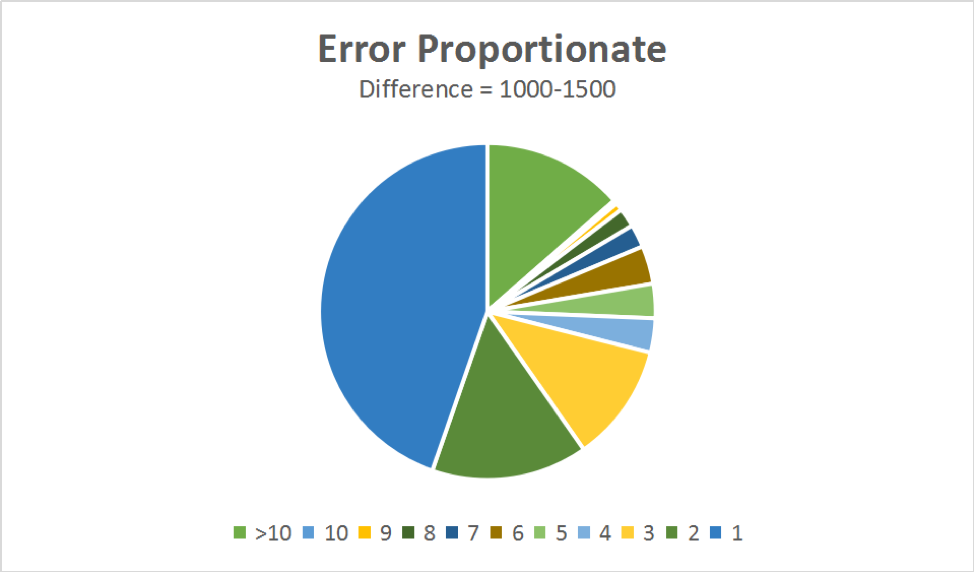}}
	\qquad
	\subfigure[difference = 1000 -1500]{
		\includegraphics[width=2.1in]{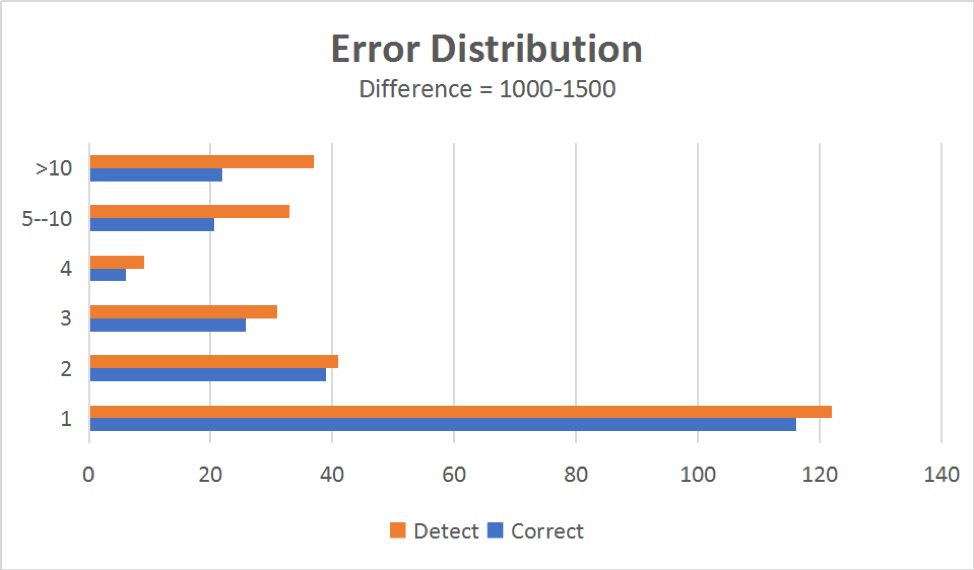}}
	\newline
	\subfigure[difference = 1500 - 2000]{
		\includegraphics[width=2.1in]{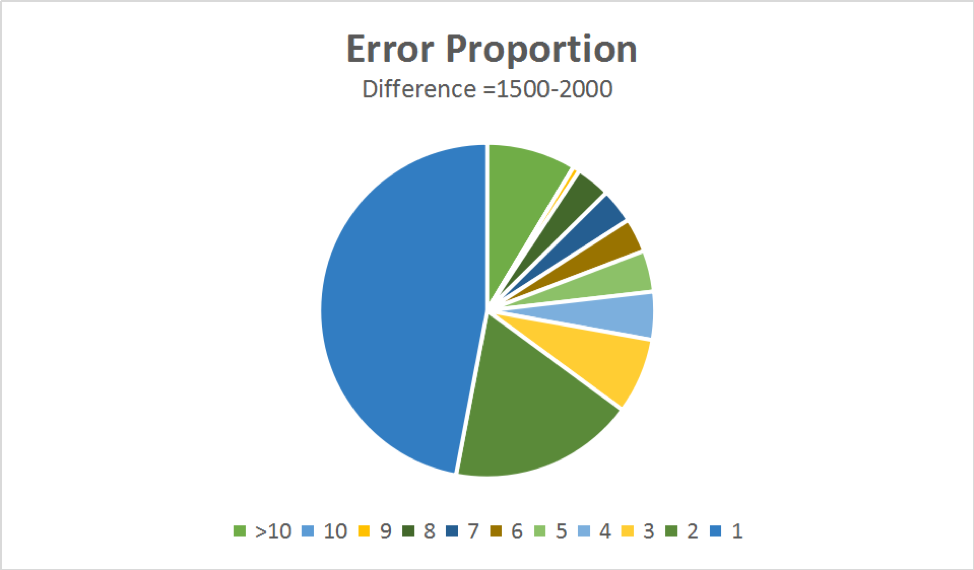}}
	\qquad
	\subfigure[difference = 1500 - 2000]{
		\includegraphics[width=2.1in]{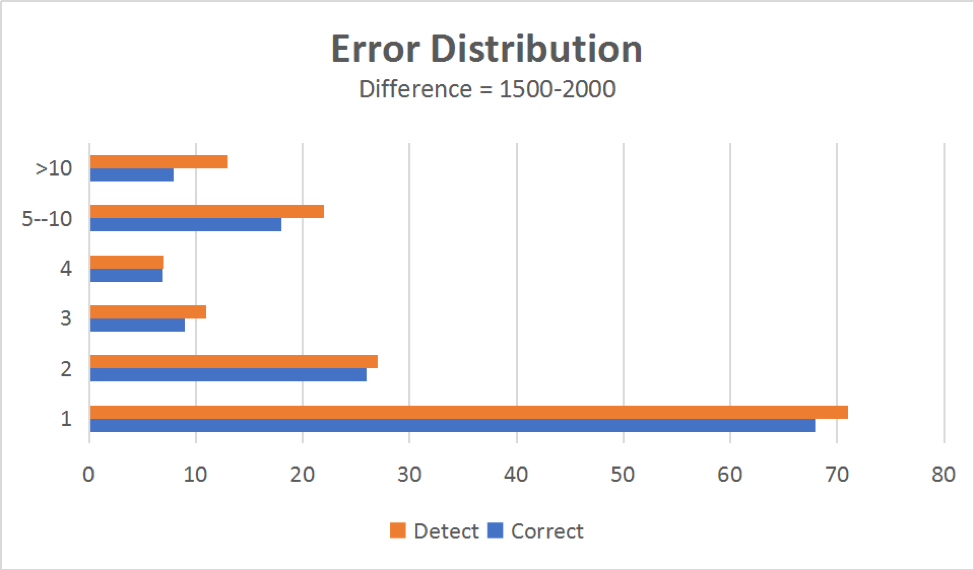}}
	\newline
	\subfigure[difference $>$ 2000]{
		\includegraphics[width=2.1in]{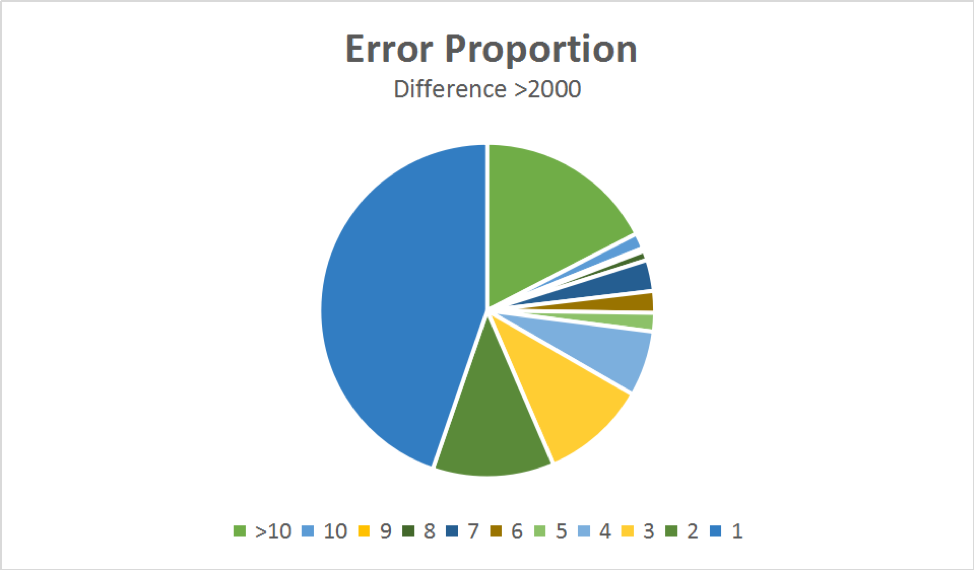}}
	\qquad
	\subfigure[difference $>$ 2000]{
		\includegraphics[width=2.1in]{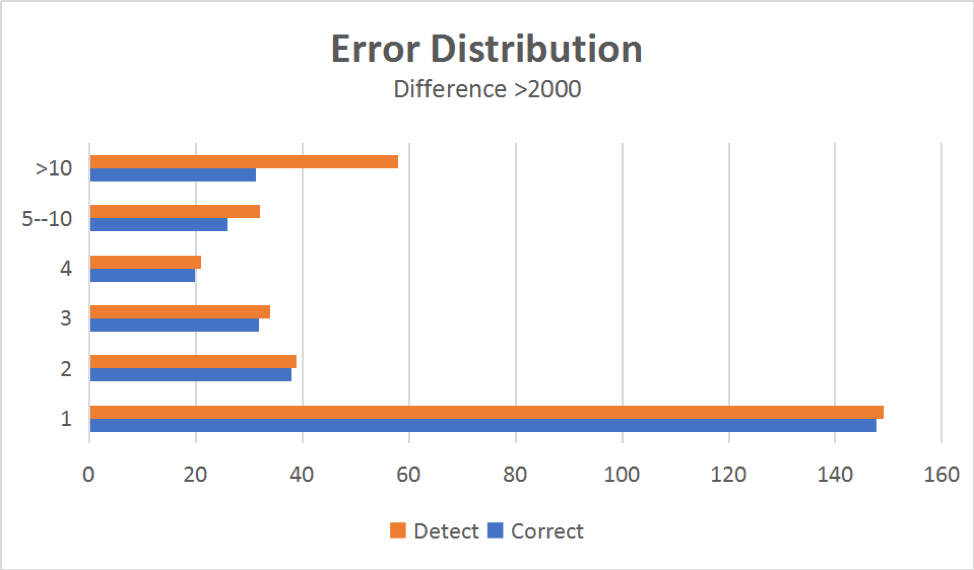}}
	\caption{Weights Influence}
	\label{fig:10fig}
\end{figure}

%------------------------------------------------

\subsection{Further Study}

There are some further experiments that we can do for a deeper examination.

\begin{itemize}
	\item To further improve the quality of knowledge bases, we are going to find better weights to classify the relations in the intersection of two conflicting concept sets.
	\item Homonyms is a hard problem in cleansing, we need to find more systematic ways to deal with it.
	\item We will further expand our method to other types of knowledge base, and provide a suggesting order on how to apply the cleansing method.
\end{itemize}

%------------------------------------------------

\section{Related Work}

\subsection{Construction and extracting of Knowledge base}
In the past, many different algorithms have been proposed for homonym extraction. Such as the simple lexical patterns~\cite{Hearst1992Automatic} , statistical and machine learning techniques~\cite{Espinosa2015Hypernym}. The usage of Isa relations are often used in induction of taxonomies. However, there are lots of errors in the automatically constructed KBs with low frequency finding in the corpus and web, and it is a big challenge to detect and repair them.

\subsection{Hash Method}
Several hash methods have been largely used on the detecting duplicate web pages and eliminating them from search results (AltaVista search engine)~\cite{Broder1997On}, and they have also been applied in large-scale clustering problems, such as clustering documents by the similarity of their sets of words~\cite{Broder2000Min}. These hash Method also gives us a technique for quickly estimating how similar the two sets are. Therefore, we could apply these methods in our research.

\subsection{Errors detecting}
Many attempts have been proposed to detect and resolve conflicts in Knowledge and databases. Li et al.~\cite{Li2004OWL} proposed  OWL- based method to detect several conflict types and resolve them in RDF knowledge base integration. Liang et al.proposed to enumerate cycles and eliminate the relation with low trustworthy score~\cite{DBLPconf/aaai/LiangXZHW17}. However, these methods do not pay attention to the properties of concept. In our paper, we propose to use the set views according to Knowledge base.

%------------------------------------------------
\section{Discussion and Conclusion}

In our work, we try to solve the problem of identifying errors of the IsA relationships from the automatically constructed Knowledge bases. Our key contribution is that we find that the intersection of two conflicting concept sets are highly likely to contain errors. We thus propose to use two different distance based models to efficiently and effectively compute the conflicting concept sets. And we analyze many influences of the error detection. Also, we suggest to give sub-attributes for homonyms and build a suspicious knowledge base for suspicious relations. We evaluate all our Models with experiment and show that our Model could produce a higher accuracy in error detecting and repairing.

\paragraph{Acknowledgements} This paper was partially supported by NSFC grant U1509216, The National Key Research and Development Program of China 2016YFB1000703, NSFC grant 61472099,61602129, National Sci-Tech Support Plan 2015BAH10F01, the Scientific Research Foundation for the Returned Overseas Chinese Scholars of Heilongjiang Provience LC2016026 and MOE-Microsoft Key Laboratory of Natural Language Processing and Speech, Harbin Institute of Technology.Hongzhi Wang is the corresponding author of this paper.

%------------------------------------------------

\section*{References}
\bibliographystyle{plain}
\bibliography{myref}

\end{spacing}
\end{document}